Extended IEEE Robotics and Automation Magazine competitions column article.

# The $10 Million ANA Avatar XPRIZE Competition Advanced Immersive Telepresence Systems

Sven Behnke, Julie A. Adams, and David Locke

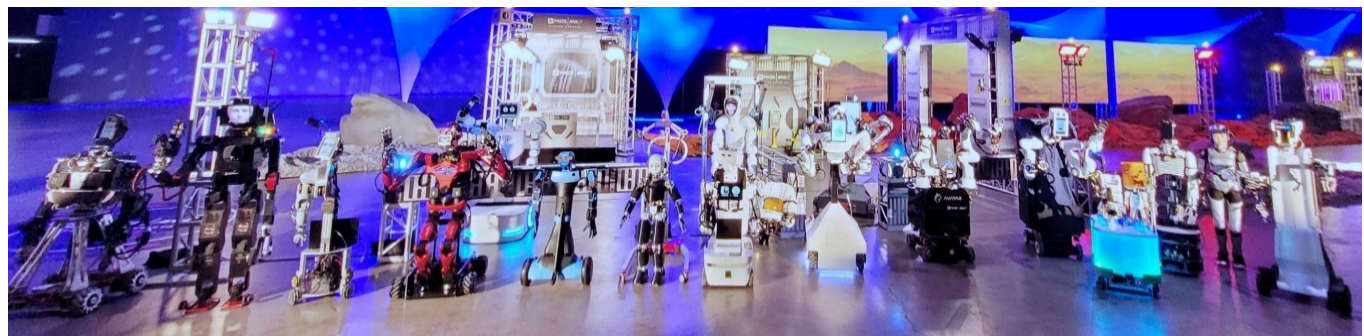

|  | AvaDynamics |  | UNIST |  | i-Botics | Tangible |  | AVATRINA |  | Pollen | Janus |
|---|---|---|---|---|---|---|---|---|---|---|---|
| Inbiodroid | Avatar-Hubo |  | SNU | AlterEgo | iCub | Cyberselves | NimbRo | Northeastern |  | Last Mile | Dragon Tree Labs |

**Fig. 1:** Avatar robots of all 17 ANA Avatar XPRIZE finals teams in front of the competition arena in Long Beach, CA.

*Abstract*— The $10M ANA Avatar XPRIZE aimed to create avatar systems that can transport human presence to remote locations in real time. The participants of this multi-year competition developed robotic systems that allow operators to see, hear, and interact with a remote environment in a way that feels as if they are truly there. On the other hand, people in the remote environment were given the impression that the operator was present inside the avatar robot. At the competition finals, held in November 2022 in Long Beach, CA, USA, the avatar systems were evaluated on their support for remotely interacting with humans, exploring new environments, and employing specialized skills. This article describes the competition stages with tasks and evaluation procedures, reports the results, presents the winning teams' approaches, and discusses lessons learned.

## I. Introduction

IMAGINE how your life would change if you could instantly transport yourself anywhere in the world. Transcending the barriers of distance and travel time would enable many activities that are impossible today, including being present at remote events, connecting with family, friends, or colleagues regardless of distance and time, and efficiently distributing skills and hands-on expertise to locations around the world wherever they are needed, e.g. to provide critical care and to deploy immediate emergency response in natural disaster scenarios. While the Star Trek transporter technology is yet to be invented, the closest thing to beaming is telepresence in an avatar robot [1].

Sponsored with 22 Million US$ by All Nippon Airways (ANA), Japan's largest airline, the Avatar XPRIZE[1] was a multi-year global competition focused on developing avatar systems that transport a human's senses, actions, and presence to a remote location in real time. XPRIZE, a nonprofit organization and global leader in designing and implementing innovative competition models to solve the world's greatest challenges, announced the Avatar competition in March 2018 with a 10 Million US$ prize purse. Fig. 2 shows the competition time line. To help shape the competition, XPRIZE formed an advisory board[2] of esteemed robotics experts, haptics inventors, psychology professors, top entertainment researchers, virtual world creators, and computer vision experts.

The participating teams had to advance and integrate multiple emerging technologies to develop a physical, non-autonomous avatar system with which an operator can see, hear, and interact within a remote environment in a manner that feels as if they are truly there. A second objective was that a person at the remote end, the recipient, would feel that the operator was present in the avatar. The avatar systems were not operated by their developers, but by members of an international expert judging panel[3] consisting of cognitive scientists, neuroscience experts, haptics authorities, wearable sensor wizards, human-robot interaction researchers, robot builders, professors, entrepreneurs, and human communication experts. Operators had only a short time to familiarize themselves with the systems before they had to perform tasks remotely in a variety of real-world scenarios, while the avatar systems had to convey a sense of presence for both the operator and the recipient in these interactions. The judges also contributed to developing the guidelines, rules, and regulations that governed the competition [2].

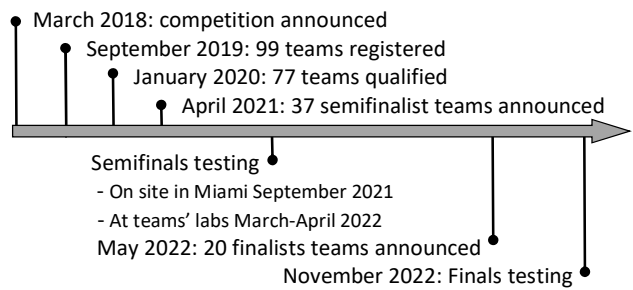

**Fig. 2:** ANA Avatar XPRIZE competition timeline.

---
[1] ANA Avatar XPRIZE https://www.xprize.org/prizes/avatar
[2] Avatar Advisory Board https://www.xprize.org/prizes/avatar/people/board

## II. QUALIFICATION

To participate, teams had to register by the end of September 2019 and submit qualifying material describing their team composition, prior work, objectives, system design approach, feasibility assessment, funding sources, and work plan by the end of October 2019. In total, 99 teams from 20 countries around the world registered. From these, an independent panel of judges selected 77 qualifying teams coming from diverse backgrounds, industries, and 19 different countries across five continents.

By February 2, 2021, qualified teams had to submit additional materials to be selected for the semifinals, consisting of a comprehensive paper detailing their avatar system's capabilities and their development plans, accompanied by a 15-minute demonstration video showing their technology completing a self-selected scenario consisting of six tasks. This gave the teams plenty of room for creativity. The addressed use-cases included remote visits and interactions with a large variety of recipients; assisting people in need by preparing food, making drinks, and helping them get dressed; and providing health care by measuring body parameters such as blood oxygen saturation. The international expert panel of judges extensively reviewed the submissions and selected 37 teams from 16 countries to participate in the semifinals. The semifinalist teams ranged from enthusiasts to university groups to leading research labs and start-up companies.

## III. SEMIFINALS

### A. Semifinal Video

As part of the semifinal verification process, the selected teams were required to submit a semifinals team video recorded in their own development workspace by July 1, 2021, along with a written description of their performance for six self-selected tasks. The video was to be up to 15 minutes long and show unedited footage from both the operator's and recipient's perspectives. Additionally, the safety system, including the emergency stop, had to be demonstrated. The addressed use-cases were similar to those in the semifinal selection video and included interaction, personal assistance, food preparation, healthcare, and building with wooden blocks.

These videos were worth ten points toward the semifinal score. Each completed task was worth one point. An additional four points were awarded for:
- the avatar robot remaining in a safe and stable position when not being actively controlled by the operator,
- the avatar robot completing the scenario without needing to be repositioned or recovered by the team members,
- the operator being able to manipulate remote objects effectively using the avatar system, and
- the form of the avatar robot being adequate for the interactions and not being intimidating or threatening.

### B. Semifinals *Testing Facilities*

The semifinals were scheduled for September 2021 in Miami, FL, USA. Because non-US teams were affected by a travel ban due to the COVID-19 pandemic, participation was a significant challenge for many qualified teams. To address this issue, XPRIZE offered semifinalists multiple testing options, including in-presence testing with support from additional U.S. team members or third-party contractors. Despite the travel ban, some non-US teams were able to participate in the semifinals with National Interest Exceptions obtained with XPRIZE's assistance or by traveling via third countries to adhere with US quarantine mandates.

In-person testing was conducted in two rounds of four days each. Semifinalist teams had two days to set up their systems and two days for test runs. Each team was given a 58 m$^2$ garage work area and a 24 m$^2$ room for setting up their operator station. Access to one of the four 45 m$^2$ avatar scenario rooms was allowed only for two-hour test slots.

The operator and scenario rooms were spatially separated and could only communicate over the XPRIZE-provided competition network with a full-duplex 1 Gbps connection. WiFi was allowed in the operator and scenario rooms to connect devices wirelessly. Communication between the operator judge and the recipient judge was only possible through the avatar systems.

Restrictions were placed on the avatar robots. Their dimensions had to be less than 100 cm wide, 120 cm long, and 210 cm high. The total weight, including batteries, had to be less than 160 kg. The avatars had to be safe for indoor use, with an emergency shutdown procedure in place and could not release any direct emissions. Drones were not allowed.

One hour of the 2 hour test slots was dedicated to training the operator, a different member of the expert jury each day, on using the team's avatar system and approaches to solving the tasks. The second hour was used for the scored test run.

### C. Semifinals Tasks

The test scenario domains were developed through extensive consultation with industry experts, the advisory board, and the judging panel. The scenarios were representative of the anticipated situations in which avatars will provide benefit to humans in the coming years. These use cases include business interactions, cultural exchanges, healthcare activities, training in various domains, and social interactions.
The three scenarios selected for the semifinal tests are shown in Fig. 3 and described in Tab. I. They capture domain aspects intended to reflect real-world situations. The test scenarios were designed to evaluate the avatar systems' ability to provide the operators with a sense of presence, as if they were actually in the remote location. The avatar systems had to transfer sensory information from the remote location back to the operator controlling the avatar.

---

[3] Avatar Judging Panel https://www.xprize.org/prizes/avatar/people/judges

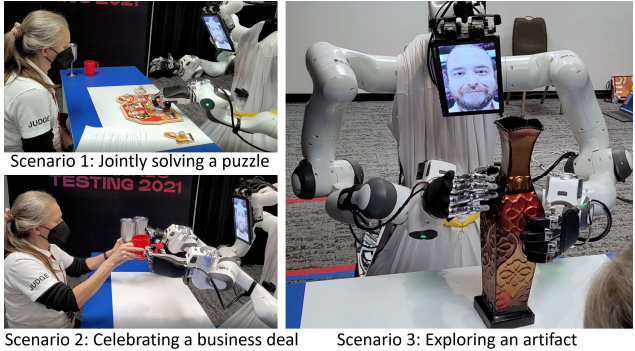

**Fig. 3:** Semifinals scenarios.

TABLE I: SEMIFINALS SCENARIOS AND TASKS
Percentages indicate fraction of points scored by the top 20 teams.

| Scenario 1: Collaborative Puzzle Activity | |
|---|---|
| Recipient R helps to finish a partially completed peg puzzle with another person (Operator O/Avatar A) at a remote location. | |
| Setup: Cleared table with simple, toddler-type puzzle with images on each piece and peg grips. R is seated on the far side of the table. O/A is positioned on the other side of the table. | |
| Task 1 100% | O/A greets R and R asks if they would like to work on the puzzle together. |
| Task 2 100% | R moves the puzzle and the detached pieces within reach of O/A and explains the process and instructions, including who goes first; the O/A verbally acknowledges that they understand. |
| Task 3 87% | O/A places a piece of the puzzle in its place while verbally identifying the image on the puzzle piece. |
| Task 4 100% | O/A points to a piece for R to do next while identifying the image on that puzzle piece. |
| Task 5 95% | O/A hands another piece to R and asks them to place the piece. This should complete the puzzle. |
| Task 6 100% | When done, the O/A and R share a gestural "high five" (with no contact) to celebrate completion of the puzzle. |
| **Scenario 2: Final Stage of a Business Deal** | |
| A business partner (O/A) is Avataring In to a meeting with a business partner acting as the host (R) at their office to celebrate the closing of an important deal. | |
| Setup: R is seated at a table with a variety of non-breakable beverage containers (coffee mugs and plastic wine glasses) set up on the table. | |
| Task 1 100% | R greets O/A. O/A waves hello. |
| Task 2 100% | O/A audibly expresses excitement about being able to meet. |
| Task 3 100% | R asks O/A which beverage they would like from a selection of beverage containers on the table. O/A chooses one beverage container by pointing to it. |
| Task 4 100% | R acknowledges the choice and moves the beverage container within range of O/A. O/A lifts the beverage container. R selects the same beverage container and lifts it. |
| Task 5 100% | While the container is lifted, O/A moves it towards Rs, and R completes a toast. The two beverage containers should touch, as in a normal toast, and are then placed back on the table. |
| Task 6 73% | R says congratulations and goodbye. O/A maneuvers away from the table to a marked, designated area. |
| **Scenario 3: A Visit to a Distant Museum of Antiquities** | |
| A visitor (O/A) is Avataring In to a Museum to explore the Museum's offerings and interact with the Museum host/greeter (R). | |
| Setup: R is seated behind a table with representative objects from the Museum on it. O/A is positioned on the other side of the table to start. Also on the table are two posters advertising different exhibits. | |
| Task 1 100% | R greets the visitor and asks O/A to indicate the exhibit they want to explore. O/A points to the desired exhibit poster. |
| Task 2 100% | R moves one artifact toward the center of the table related to that exhibit and asks O/A to describe it. O/A describes the artifact visually. |
| Task 3 100% | R says the O/A can touch the artifact. O/A then explores the artifact by touching it. |
| Task 4 73% | O/A explores the object and describes the texture they feel. |
| Task 5 67% | O/A picks up the artifact and describes the weight of the object and places it back on the table. |
| Task 6 67% | R points to where O/A can find the start of their exhibit marked on the floor. O/A maneuvers away from the table to the marked, designated area. |

This sensory information included aspects of the location, such as understanding the layout, placement of objects, path navigation, and other location-specific cues. The tasks also included interactions with the recipient in the remote location through recognition, understanding, communication, gestures, and shared experiences. Teams were required to attempt all three scenarios. Re-attempts were possible at the request of the teams or the operator judges.

*D. Semifinals Scoring*

Each scenario was worth up to 30 points. The judges evaluated the avatar systems based on four categories: operator experience (12 points), recipient experience (8 points), avatar ability (6 points), and overall system (4 points), which are detailed in Tab. II.

For each scenario, the better score of the two competition days was kept. As evident from the score weights, the semifinal testing focused heavily on the subjective experience of the operator feeling present in the remote room and the recipient perceiving the operator's actions through the robotic avatar. Less emphasis was placed on mobility, task completion, and system reliability.

TABLE II: SEMIFINALS SCORING
Percentages indicate the fraction of points scored by the top 20 teams.

| Operator Experience (12 points) |
|---|
| - O felt present in the remote space with R. 87% |
| - O was able to sense or understand R's emotion. 92% |
| - O was able to express their emotions to R. 86% |
| - O was able to clearly see and hear what was happening in the remote space. 77% |
| - O was able to get the necessary tactile/haptic/force/other feedback to complete the required tasks. 54% |
| - O was able to sense their own position and movements in the remote space. 79% |
| - O was able to move around in the remote space to complete the required tasks. 82% |
| - O was able to manipulate remote objects effectively. 80% |
| - O felt they were able to gesture effectively to R. 89% |
| - O felt safe using the Avatar System. 90% |
| - O felt the Avatar System was easy to use. 84% |
| - O felt the Avatar System was comfortable to use. 84% |
| **Recipient Experience (8 points)** |
| - R was able to identify O and felt O was present in the space. 89% |
| - R was able to understand O's communications. 94% |
| - R was able to understand O's emotions. 82% |
| - R felt a sense of shared experience with O. 85% |
| - R felt safe while A was navigating the environment and manipulating objects. 86% |
| - R felt A's aesthetics were adequate to the interactions and not intimidating or threatening. 85% |
| - R was able to understand O's gestures. 85% |
| - R felt O could understand them. 98% |
| **Avatar Ability (6 points)** |
| Each task listed in Tab. I scored one point when completed. |
| **Overall System (4 points)** |
| Scored by O: |
| - Did the avatar system (hardware and power) operate reliably enough for O to attempt all the tasks? 100% |
| - Did the avatar system (software and network) operate reliably enough for O to attempt all the tasks? 98% |
| Scored by R: |
| - Did A remain in a safe and stable position when not being actively controlled by O? 98% |
| - Did A complete the scenario without needing to be repositioned or recovered by the team? 98% |

*E. Semifinals Results*

Twenty-nine teams participated in the semifinals event in Miami. Six additional teams that were unable to travel to Miami were visited by XPRIZE personnel and judges, who conducted testing in their own development workspaces from March to April 2022. The developed avatar systems varied widely in complexity and capability. Avatar robots had one or two manipulator arms with grippers of varying dexterity: from simple parallel grippers to human-like five-finger hands. The avatars often had a movable head with cameras and microphones. Some teams also displayed the operator's face or animated their facial expressions in the avatar's head [3]. Most avatar robots moved on wheels, but some used two legs for walking. Most operator stations used head-mounted displays to provide the visualization. Often, the operator used handheld VR controllers or data gloves to teleoperate the avatar. Some teams captured operator inputs via upper-body and hand exoskeletons, providing the operator with force and haptic feedback.

A total of 214 scored scenarios were completed. 34 teams completed at least one run and 31 teams completed all six scenarios. There were 12 reruns (only 5 with the top 20 teams).

Table III lists the top 20 teams that qualified for the finals, ranked by score. 2 Million US$ prize money was paid in equal portions to the top 15 teams that tested in Miami. The first-ranked team NimbRo [4] scored 99/100 points, losing only one point in Scenario 1. Tables I and II report the percentages of points scored by the top-20 teams for the individual tasks and for the questions answered by the operator and recipient judges, respectively. There were 13 different operator judges and 12 different recipient judges involved. One can observe that most of the experience scores were very high. The lowest scores indicate the difficulty for the operator to feel haptics and forces remotely (54% of points scored), to see and hear clearly what was going on in the remote space (77%), and to sense their own position and movements in the remote space (79%). Consequently, the most difficult tasks involved estimating the weight of an object (S3T5), haptic perception (S3T6), and mobility (S2T6, S3T6).

TABLE III: SEMIFINALS RESULTS

| Rank | Team Name | Country | Tested | Score |
|---|---|---|---|---|
| 1 | NimbRo [4] | Germany | Miami | 99 |
| 2 | iCub [5] | Italy | own lab | 95.25 |
| 3 | i-Botics [6] | Netherlands | own lab | 93.75 |
| 4 | T. Northeastern [7] | Unites States | Miami | 93 |
| 5 | Dragon Tree Labs[4] | Singapore | Miami | 93 |
| 6 | AVATRINA [8] | United States | Miami | 92.75 |
| 7 | Avatar Hubo [9] | United States | Miami | 92 |
| 8 | Tangible[5] | United States | Miami | 92 |
| 9 | AlterEgo [10] | Italy | own lab | 91.75 |
| 10 | Cyberselves[6] * | Un. Kingdom | Miami | 90.75 |
| 11 | Team SNU [11] | South Korea | Miami | 89.5 |
| 12 | Pollen Robotics[7] | France | Miami | 89,5 |
| 13 | Last Mile[8] | Japan | Miami | 88.5 |
| 14 | Enzo† | Colombia | own lab | 87.25 |
| 15 | Team UNIST[9] | South Korea | Miami | 86 |
| 16 | Inbiodroid[10] | Mexico | Miami | 84.5 |
| 17 | Rezillient[11] † | United States | Miami | 84 |
| 18 | Touchlab[12] * | Un. Kingdom | Miami | 82.5 |
| 19 | AvaDynamics[13] | United States | Miami | 80.5 |
| 20 | Janus [12] | France/Japan | own lab | 80 |

* merged for Finals. † dropped out from competition prior to Finals.

---

[4] Dragon Tree Labs https://www.dragontreelabs.tech
[5] Converge Robotics https://www.convergerobotics.com
[6] Cyberselves https://www.cyberselves.com
[7] Pollen Robotics https://www.pollen-robotics.com
[8] Team Last Mile http://ogilab.kutc.kansai-u.ac.jp/small_world/indexorg
[9] UNIST BiRC Lab http://birc.unist.ac.kr
[10] Inbiodroid https://inbiodroid.com
[11] Rezilient Health https://www.rezilienthealth.com/ana-avatar-xprize
[12] Touchlab https://www.touchlab.io
[13] Team AvaDynamics http://avadynamics.com

## IV. Finals

The finals were held November 1-5, 2022 at the Long Beach Convention & Entertainment Center, Los Angeles County, CA, USA. Two qualified teams merged and two dropped out, leaving 17 teams from 10 countries, distributed as follows: North America (6 teams), Europe (6 teams), and Asia (5 teams). All the avatar robots participating in the finals are shown in Fig. 1. Team size varied from 4 to 19 members, with an average of 12 members, for a total of 200. 10 teams were from universities or research labs. The remaining teams were from start-ups, small technology-driven companies, and enthusiasts.

Parallel to the finals, a technology fair was held with 26 booths from companies and research institutes exhibiting teleoperated robotic systems and components.

### A. Finals Testing Facilities

Teams were given $37\,m^2$ of garage space and two days to set-up their systems. The avatar robots were tested on a single test track of $240\,m^2$, shown in Fig. 1. Five operator control rooms of approximately $24\,m^2$ were shared by multiple teams; thus, an additional requirement for the finals was that the operator station had to be moved in and out of these rooms and set up in 30 minutes. Furthermore, the avatar robots had to be untethered, meaning that they were powered by batteries and communicating only via the XPRIZE-provided WiFi. Team members were not permitted on the test course and gantries were only permitted as part of the robot itself.

### B. Finals Mission and Tasks

The finals testing required the avatar to complete a mission that consisted of ten tasks, described in Table IV, resembling a mission on a distant planet. The tasks were grouped into three domains: connectivity, exploration, and skill transfer. Tasks had to be completed in the specified order and included navigation among obstacles, multimodal communication with a human recipient, simple and complex object manipulation, judging the weight of objects, using a power drill, and discriminating texture, or roughness of stones. Fig. 4 shows these tasks. The maximum mission time was 25 minutes and completion of all tasks was required to be eligible for the grand prize.

### C. Finals Scoring

The 18 members of the expert jury had five distinct roles:
- Operator Judge (the avatar): Operated the avatar systems during testing runs. Teams had 45 minutes in the control room to train the assigned operator judge on how to use their system. The operator then used the system to conduct the test run that had a maximum time limit of 25 minutes. The operator judge scored three subjective presence questions.
- Recipient Judge (mission commander): Interacted with the avatar system on the test course and provided direction for accomplishing the mission. The recipient judge scored two subjective presence questions.
- Official Scorer (mission control): Monitored the live testing run, ensured all tasks were completed and objectively awarded points.

TABLE IV: Finals Tasks and Scoring Questions
The percentages report task completion rates of the 12 best teams.

| | |
|---|---|
| **Connectivity: Human-to-human connection** | |
| Task 1 100% | The avatar robot maneuvers to the mission control desk. Was the avatar able to move to the designated area? |
| Task 2 100% | The Avatar reports to the mission commander and introduces themselves. Did the avatar introduce themselves to the mission commander? |
| Task 3 100% | The avatar receives the mission details and confirms them with the mission commander. Was the avatar able to confirm (repeat back) the mission goals? |
| Task 4 100% | The avatar activates a switch which opens the station door. Was the avatar able to activate the switch? |
| **Exploration: The new era of travel** | |
| Task 5 96% | The avatar exits the mission control room through the door and travels across the planet to the next task. Was the avatar able to move to the next designated area? |
| Task 6 79% | The avatar must identify the full power canisters that are among empty canisters. Was the avatar able to identify the heavy canister? |
| Task 7 71% | The avatar places the correct canister into the designated slot which triggers the lighting of the next task zone. Was the avatar able to lift up and place the heavy canister into the designated slot? |
| **Skills Transfer: Expertise with no boundaries** | |
| Task 8 63% | The avatar navigates along the planet's surface to arrive at the next task. Was the avatar able to navigate through a narrow pathway to get to the designated area? |
| Task 9 42% | The avatar must use the drill to remove the door. Was the avatar able to utilize a drill within the domain area? |
| Task 10 25% | The avatar must reach through the barrier to identify the rough textured rock and retrieve it. Was the avatar able to feel the texture of the object without seeing it, and retrieve the requested one? |

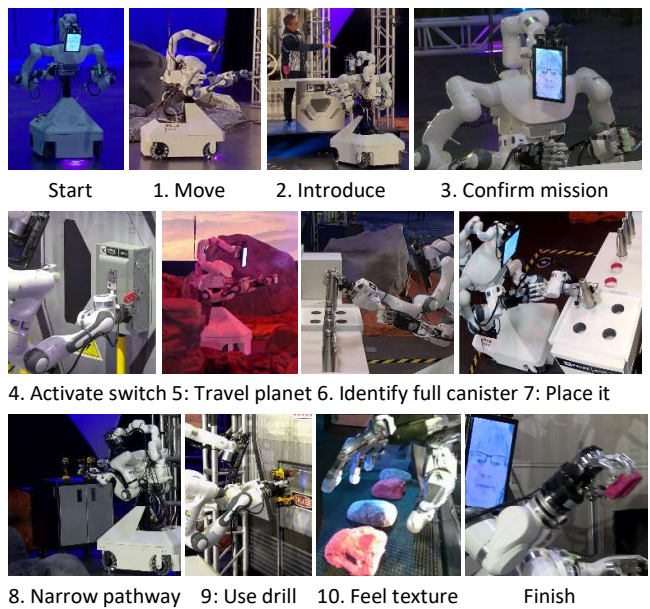

**Fig. 4:** Finals tasks, see descriptions in Tab. IV.

- Staging Area Coordinator: Coordinated with competing teams preparing their avatar to go onto course.
- Test Course Manager: Ensured that the test course was set up identically between runs, that teams' avatars performed

TABLE V: FINALS EXPERIENCE SCORING

| Operator Experience (3 points) |
| --- |
| - The avatar system enabled the operator judge to feel present in the remote space and conveyed appropriate sensory information. |
| - The avatar system enabled the operator judge to clearly understand (both see and hear) the recipient. |
| - The avatar system was easy and comfortable to use. |
| **Recipient Experience (2 points)** |
| - The avatar robot enabled the recipient judge to feel as though the remote operator was present in the space. |
| - The avatar robot enabled the recipient judge to clearly understand (both see and hear) the operator. |

TABLE VI: FINALS RESULTS

| Rank | Team | Time | Task | Judged | Total |
| --- | --- | --- | --- | --- | --- |
| 1 | NimbRo (DE) | 5:50 | 10 | 5 | 15 |
| 2 | Pollen Robotics (FR) | 10:50 | 10 | 5 | 15 |
| 3 | Team Northeastern (US) | 21:09 | 10 | 4.5 | 14.5 |
| 4 | AVATRINA (US) | 24:47 | 10 | 4.5 | 14.5 |
| 5 | i-Botics (NL) | 25:00 | 9 | 5 | 14 |
| 6 | Team UNIST (KR) | 25:00 | 9 | 4.5 | 13.5 |
| 7 | Inbiodroid (MX) | 25:00 | 8 | 5 | 13 |
| 8 | Team SNU (KR) | 25:00 | 8 | 4.5 | 12.5 |
| 9 | AlterEgo (IT) | 25:00 | 8 | 4.5 | 12.5 |
| 10 | Dragon Tree Labs (SG) | 25:00 | 7 | 4 | 11 |
| 11 | Avatar Hubo (US) | 25:00 | 6 | 3.5 | 9.5 |
| 12 | Last Mile (JP) | 25:00 | 5 | 4 | 9 |

the test mission safely, and that no team members entered the test track during the trial period.

Completion of each task was worth one point. Operator and recipient judges scored their experience of being present in the remote space and of perceiving the presence of the remote operator, respectively, using the criteria listed in Tab. V. Each criterion was assessed on a scale of never/poor (0 points), sometimes/fair (0.5 points), and always/good (1 point). Thus, up to five experience points could be earned.

*D. Finals Results*

Finals testing began with a qualifying day. All teams had their systems operated by judges on the test course and were scored as described. The top 16 teams (and ties based on points) receiving a minimum of 4 points advanced.

The two finals testing days were open to the public, with more than 2,100 visitors and many media representatives in attendance [13] and thousands watching the live stream[14]. This was the first time since the Ansari XPRIZE in 2004 that XPRIZE conducted a public testing event. The test runs were presented by a professional commentator, a technical analyst, and a field reporter who interviewed team representatives, judges, and experts. Pre-recorded material was used to fill in gaps.

All 17 teams were tested on Test Day 1, 16 of which scored 4-10 task points and 3-5 experience points, for a total of 7-15 points. Three teams completed all ten tasks, with NimbRo in the lead (15 points), followed by Pollen Robotics (14.5 points), and AVATRINA (14.5 points). The top 12 teams advanced and were tested on Day 2. Table VI reports the final results. The better of the two scores was kept, and the completion time was used to break ties.

The task completion rates reported in Tab. IV indicate that the top 12 teams solved all four tasks in the connectivity domain on both days. The first task that multiple teams failed to complete was Task 6, which involved discriminating between objects of different weight. The drop in completion rates for Tasks 8-10 was caused by two factors: lack of time and task difficulty. Specifically, grasping and using the power drill to remove a screw was successful only 67% of the time, and identifying a rough stone by feeling its texture and retrieving it had only 60% success rate.

The two teams that completed all tasks on both days improved their execution time on Day 2 by 29% and 11%, respectively.

The experience scores given by the operator and recipient judges were high, with an average score of 4.5 points, a minimum score of 3.5, and four teams receiving a perfect score of 5. This result indicates that, overall, the judges were quite satisfied with their experience. The recipient judges rarely deducted half a point, but the operator judges deducted points significantly more frequently.

Four teams completed all ten tasks, with two receiving perfect experience scores from both judges. The fastest team, NimbRo, took an average of only 35 seconds per task, which is not much slower than a human performing the tasks directly would need. The top three teams received a prize of $5 million (NimbRo), $2 million (Pollen Robotics), and $1 million (Team Northeastern).

*E. Finals Observations*

Overall, finals testing worked well and the developed avatar systems were properly evaluated. This was made possible not only by XPRIZE, but also by the hard work of the international panel of expert judges, who operated the avatar systems, acted as recipients, and administered and scored the tests.

The tight schedule of finals testing in a single competition arena with few operator control rooms was a significant challenge for both XPRIZE and the teams. Especially on Qualification Day, when the avatar systems were tested in the competition arena for the first time, everyone had to adjust to the setting, which caused some delays. With practice, testing went more smoothly on Day 1 and was on schedule on Day 2.

Overall avatar system reliability was an issue, though. In 32% of the test runs, it took a significant amount of time for the avatars to start moving. There were also several stoppages and delays during the tests. Of course, the avatar systems being tested were complex and had many components that could fail. The largest single cause of failure was network connectivity issues, but cable disconnections, software freezes, etc. also occurred.

Another issue for some teams was the operator's situational awareness. For example, it was not always easy for the operator to choose an appropriate communication distance to the recipient. Some avatar robots collided with test course objects and obstacles, sometimes resulting in falls. In terms of mobility, wheeled robots had a clear advantage over walking robots, for which balance was a challenge.

---

[14] Finals live stream recording: https://youtu.be/lOnV1Go6Op0

Judges sometimes struggled to remember how to use the systems. The operators quickly became immersed in the remote space, using the avatar robot as their own body, but if they had to communicate with team members in the operator room, that immersion broke.

Force and haptic feedback were necessary for Tasks 6 and 10 to identify a heavy object and to feel the texture of stones, but was also useful for other tasks and strongly enhanced the feeling of presence.

Two arms were not required for any task, but were often helpful, e.g. to hold two canisters simultaneously to compare their weights, or to pick up or fix an object with one hand to grasp or scratch it with the other. Two arms were also often equipped with different grippers, providing complementary capabilities for different tasks and also provided some redundancy in case one arm failed.

The ability for operators to change their viewing perspective without moving the robot base was important, as it increased immersion and 3D scene perception, minimized occlusion, and allowed for selecting appropriate views for manipulation tasks.

Some form of animation or display of the operator's face on the avatar robot helped the recipient judge to experience the operator's presence and to see them clearly.

### IV. Developed Avatar Systems

A large number of capable avatar systems were developed for this competition. Importantly, advances in the state of the art of telepresence and mobile telemanipulation in the avatar robots, the operator interfaces, and the communication between the two were achieved. We exemplarily present the approaches of the three winning teams.

*A. NimbRo*

NimbRo is the robot competition team of the Autonomous Intelligent Systems Lab at the University of Bonn, Germany.

The NimbRo avatar robot (Fig. 5 right) has a human-like upper body with two compliant 7 DoF arms (Franka Emika Panda, 3 kg payload, 85.5 cm reach) equipped with dexterous five-finger hands. The right hand (Schunk SVH) is very human-like in proportions and dexterity with 20 DoF actuated by nine motors, while the cable-driven left hand (Schunk SIH) is larger, more compliant and has only 11 DoF, actuated by five motors. For haptic perception, the SIH fingertips are equipped with 3D Hall effect sensors. This hand's index finger is also equipped with two different microphones (piezo-based contact microphone and MEMS microphone). The SVH fingertips are equipped with miniature switches. The hands are attached via 6-axis force/torque sensors.

The robot's head is mounted on a 6 DoF (UFactory xArm) robotic arm that provides full freedom of movement, mirroring the operator's head motion. The head consists of a display showing the animated operator's face, a pair of wide-angle cameras (Basler) with a human-like baseline providing

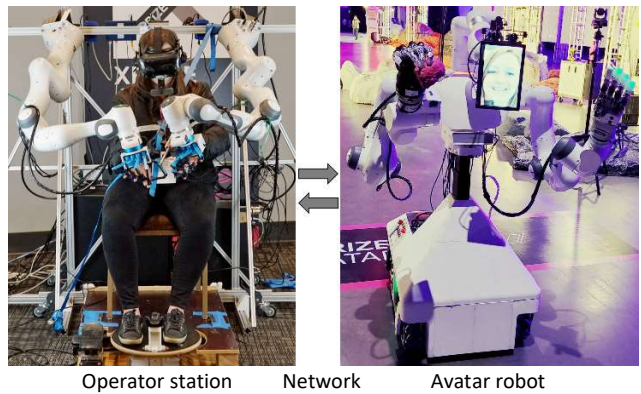

Operator station    Network    Avatar robot

**Fig. 5:** Avatar system of winning team NimbRo.

4K video streaming at 46 Hz, and a stereo microphone.

The upper body is attached to the robot base via a linear actuator, allowing manipulation at various heights, including the floor. Four Mecanum wheels provide omnidirectional driving capabilities. Two wide-angle cameras mounted on the front and the back of the robot's torso provide a birds-eye view around the robot while driving, facilitating safe obstacle avoidance. The robot is powered by a 48V 30Ah battery and has a fast on-board PC with Nvidia GPU and two WiFi adapters.

The operator station (Fig. 5 left) is equipped with two 7 DoF arms (Franka Emika Panda) and 20 DoF hand exoskeletons (SenseGlove DK1) that measure the operator's arm and finger movements. The hand exoskeletons are attached via 6-axis force/torque sensors. The arms provide force feedback to the operator's hands [14] and the exoskeletons exert resistance to finger closure via brakes. Further haptic feedback is provided to the fingers by vibrating actuators. The left index finger is also equipped with a linear actuator for continuous force feedback and a strong vibration actuator for haptic feedback [15].

The operator wears a VR head-mounted display (HTC Valve Index) with stereo headphones, allowing full immersion in the remote situation (Fig. 6). Spherical rendering [16] is used to mitigate latency of the camera motion, preventing motion sickness. The HMD has been equipped with three additional cameras to capture the operator's mouth and eyes. Based on this, the operator's gaze direction and eyelid opening are estimated in order to photorealistically animate their face in real time [17]. The operator can control the omnidirectional driving with a 3 DoF foot controller and upper body height with a pedal.

Separate ROS cores run on the operator station and the avatar robot. They communicate bidirectionally via UDP using the nimbro_network library[15]. The main camera stream (stereo 2472×2178 @46 fps) is HEVC-encoded and decoded on the GPU (NVENC), resulting in a bandwidth of 14 Mbps. Audio is processed by the JACK Audio Connection Kit and redundantly transmitted via UDP using the OPUS audio codec. NVIDIA MAXINE is used for GPU-accelerated acoustic echo cancellation and Jamulus for team communications with operator and recipient. Control data is sent redundantly and packet loss is monitored. System operation is constantly monitored and components are auto-respawned when necessary.

---

[15] NimbRo network library: https://github.com/AIS-Bonn/nimbro_network

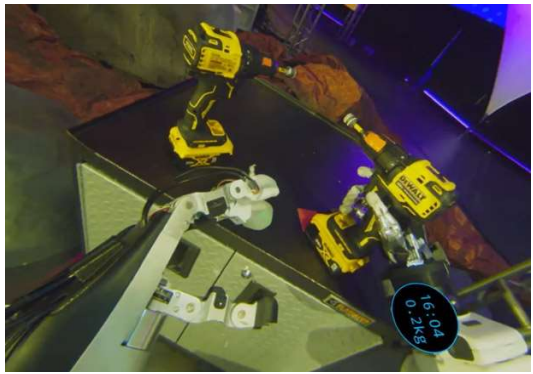

**Fig. 6:** Immersive visualization of grasping a power drill.

More detail on NimbRo's avatar system is available in [18].

*B. Pollen Robotics*

Pollen Robotics SAS is a telepresence startup based in Bordeaux, France. Their avatar robot (Fig. 7 left) is a version of their telepresence robot Reachy[16] that was improved in several aspects for the competition. The upper body is human-like with two 7 DoF arms, each with a payload capacity of 3.5 kg. The arms are driven by self-designed parallel Orbita actuators: 2 DoF in the shoulder and elbow and 3 DoF in the wrist. The grippers have three underactuated fingers that adapt to objects. One finger is equipped with a fingernail and a microphone to provide haptic feedback to the operator. The robot head consists of a facial animation display, a stereo camera (ZED mini), and two microphones. It is attached to the torso via a 3 DoF Orbita actuator. The robot drives using a three-wheeled omnidirectional base with a compact, circular footprint that is equipped with a 2D horizontal LiDAR sensor to provide the operator with a navigation map.

The operator station provides immersive audio-visual feedback through an HMD (Oculus Quest 2) equipped with a facial tracker (VIVE) to capture the operator's mouth area for animating the avatar's face. The Oculus VR controllers used to capture the operator's hand movements were augmented with voice-coil vibration actuators (Actronica) to provide haptic feedback. The mobile base is controlled via a joystick on one of the controllers. 1 DoF actuators provide torque feedback to the operator's elbows. WebRTC is used for bi-directional communication between the avatar robot and the operator station.

*C. Team Northeastern*

Team Northeastern is based at the Institute for Experiential Robotics at Northeastern University, Boston, MA, USA. Their avatar robot Robalto (Fig. 7 right) is equipped with two 7 DoF arms (Franka Emika Panda) in a non-anthropomorphic configuration. The self-designed grippers have three fingers with low-impedance hydrostatic transmissions that provide haptic force feedback to the operator. The low-fric-tion finger actuators and hydraulics with pressure sensors are located in the robot base [19]. A display in the robot's center shows the operator. The robot is equipped with two wide-angle cameras: one main camera for manipulation and a top-down camera for navigation. The robot base (MassRobotics) drives omnidirectionally using four Mecanum wheels.

[16] Pollen Robotics Reachy https://www.pollen-robotics.com/reachy

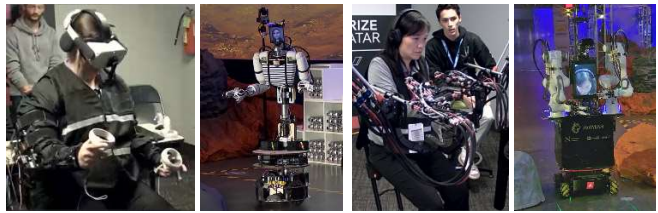

**Fig. 7:** Avatar systems of 2nd place team Pollen Robotics (left) and 3rd place Team Northeastern (right).

The operator station provides force feedback to the operator's hands using two 3 DoF exo-arms. The operator wears hand exoskeletons with gimbals to track hand and finger motions and to provide force feedback to the fingers. Avatar driving is controlled with an omni-directional pad.

The operator sees the remote scene on two large screens: one in portrait mode showing the manipulation workspace and a wide-angle landscape screen on the floor showing a top-down view for navigation. The avatar projects two parallel laser lines in the forward direction onto the floor, indicating its width to the operator for navigation, which is necessary to compensate for the lack of a 3D visualization. Similarly, two lateral laser lines tracking the gripper's forward motion indicate depth for manipulation.

More detail on Team Northeastern's avatar system is available in [20].

## VI. DISCUSSION

The ANA Avatar XPRIZE competition was a multi-year collaborative effort by the organizers, judges, and participating teams to advance the state of the art in telepresence robotics. The competition was well-organized with multiple milestones and team down-selections. The competition's focus changed along the way and, hence, it was also evaluating the team's abilities to adapt to unexpected requirements. During the semifinals, the focus was on the interaction between the operators and the recipients, and the judges enjoyed testing the different avatar systems' capabilities. Teams could define their own tasks and demonstrated them in a video that contributed to the score. In contrast, the finals focused more on capabilities, task completion, and overall system reliability. Since execution time was used for scoring, judges had no time to playfully explore an avatar's capabilities except during operator training and after the competition.

When designing a competition's rules and tasks, it is not easy to define the difficulty level. The competition tasks need to be ambitious, but should also be achievable by the best teams. XPRIZE and the judges struck the right balance, as evidenced by the competition results. The clear scoring criteria and judge rotations resulted in a fair and reproduceable evaluation of the developed avatar systems, even for subjective criteria, such as the feeling of being present in the remote space.

The developed avatar systems significantly advanced the field, including the immersive visualization of a remote scene, force and haptic feedback to feel the interaction with the environment, and animation of the operators' faces by the avatar. Head movements, gaze, gestures, and facial animations contributed to perceiving the operator being in the avatar and establishing shared attention with the recipient. It remains open, however, how much human-likeness is necessary or desirable and what form avatars should have.

Rather than converging to a single solution, the competition featured a variety of different approaches that performed well. Success was determined not only by the technical capabilities of the robotic systems, but also by the intuitiveness of the operator interfaces and the effectiveness of the operator training procedures.

The largest prize purse in a robotic competition to date attracted top research groups and companies. While teams had different policies regarding the disclosure of their technical approaches and results during the competition, XPRIZE organized a workshop after the finals to share developments and discuss lessons learned. This workshop was followed-up by a session at the 2022 IEEE Telepresence Symposium and the 2nd Workshop Toward Robot Avatars at the 2023 IEEE International Conference on Robotics and Automation.

After the competition, participants moved on to new challenges. Some teams wish to commercialize the developed technologies, which requires identifying viable use cases. The complex avatar systems could be further developed for dangerous or hard-to-reach domains such as space, disaster relief, or medical assistance in isolation wards. Everyday virtual travel use requires avatar systems to become simpler and more affordable. Academic groups will explore research questions raised by the competition, including how much human-likeness avatars should assume and how to balance and interface direct control and autonomy.

XPRIZE is discussing a follow-up avatar competition with potential sponsors. This potential follow-up could raise the bar in several dimensions, including imposing bandwidth restrictions and latencies, locomotion on more difficult terrain, more complex manipulation (e.g., bimanual tasks), additional interaction modalities (e.g., temperature or smell), and deeper interactions between avatars and recipients including interpretation of subtle communication cues and direct physical contact.